\documentclass{bmvc2k}
\usepackage{siunitx}
\usepackage{gensymb}
\usepackage{amssymb}
\usepackage{booktabs}
\usepackage[table]{xcolor}
\usepackage{subcaption}
\usepackage{pifont}
\newcommand{\cmark}{\ding{51}}
\newcommand{\xmark}{\ding{55}}
\usepackage{multirow}
\usepackage{enumitem}

\title{TALON: Token-Aligned Lightweight Adapters for 6-DoF Spacecraft Pose Estimation}

\addauthor{Abid Ali}{}{} %{1}
\addauthor{Nidhal Eddine Chenni}{}{}
\addauthor{Arunkumar Rathinam}{}{}%{1}
\addauthor{Djamila Aouada}{}{}%{2}

\addinstitution{
$CVI^2$ SnT, University of Luxembourg
}
\runninghead{Student, Prof, Collaborator}{BMVC Author Guidelines}

\begin{document}

\maketitle

\begin{abstract}
Monocular 6-DoF spacecraft pose estimation methods predominantly process frames independently, discarding the temporal structure present in approach and inspection manoeuvres. The few temporal approaches require full backbone fine-tuning or auxiliary optical flow networks, risking catastrophic forgetting and increasing computational cost. We propose TALON (Token-Aligned Lightweight adapters for Orbital Navigation): factorised spatiotemporal 3D adapters injected before the self-attention layers of a frozen DINOv3 vision transformer, combined with a patch-token alignment loss that geometrically grounds the adapted features to keypoint structure through a prototype-conditioned KL-divergence objective. Pre-attention placement allows the frozen attention to reason over temporally enriched tokens, achieving stronger performance with a single adapter per block than post-attention alternatives. The alignment loss shapes the intermediate representations so that each keypoint induces a spatially precise activation in the token field, improving fine-grained localisation (PCK@1: $3.3{\times}$ over existing temporal methods) while adding less than 5\% parameters to the frozen backbone. On SPADES, TALON achieves ${E}_{R}{=}0.0910$\,rad with a 31\% reduction in pose score over the prior state-of-the-art. On the SwissCube benchmark, TALON attains 78.8\% ADD-0.1d accuracy, surpassing the prior best by 21.8\%. Cross-domain generalisation on SPARK reduces pose error by $2.5{\times}$, and ablations reveal adapter depth as a control for the trade-off between specificity and generalisation across domains.
Github link: \href{https://github.com/abalikhan/Token-aligned_adapter}{https://github.com/abalikhan/Token-aligned\_adapter}
\end{abstract}

%-------------------------------------------------------------------------

\section{Introduction}
\label{sec:intro}

Estimating the six degree-of-freedom (6-DoF) pose of a noncooperative spacecraft from monocular imagery is a core capability for autonomous rendezvous, proximity operations, on-orbit servicing, and active debris removal\cite{opromolla2017review, pauly2023survey}. Deep learning has driven rapid progress on this task, with both direct regression and hybrid keypoint-based pipelines achieving strong results on benchmark datasets\cite{sharma2018pose, park2019pix2pose, chen2019satellite, hu2021wide}. Vision foundation models such as DINOv2\cite{oquab2024dinov2} and DINOv3\cite{Simeoni2025DINOv3} produce patch-level features that encode rich semantic and spatial information, and recent work in general 6-DoF pose estimation has shown that these frozen features can be used directly for template matching and correspondence\cite{caraffa2024freeze, ornek2023foundpose}. However, a fundamental gap remains: foundation model patch tokens are trained to be general-purpose descriptors, not to encode the specific geometric keypoint structure of a target object. Bridging this gap without destroying the pre-trained representations through full fine-tuning is the central problem addressed in this work.

This problem is compounded in the spacecraft domain by two challenges. First, most spacecraft pose estimation (SPE) methods process each frame independently\cite{park2022speed+, ancey2025fastposevit}, discarding sequential structure and yielding temporally inconsistent pose trajectories. The few methods that incorporate temporal information either require full fine-tuning of large backbones\cite{rondao2022chinet, musallam2021leveraging} or rely on auxiliary networks such as optical flow estimators\cite{sosa2025motionaware, zuo2024crospace6d}, substantially increasing computational cost. Second, annotated spacecraft data remains scarce, making parameter-efficient adaptation essential: full fine-tuning risks catastrophic forgetting of the general representations that make foundation models valuable in the first place.

We address these challenges with a unified framework built around a single principle: the patch tokens of a frozen foundation model should be shaped to simultaneously encode temporal context and geometric keypoint structure, without altering the pre-trained backbone. To inject temporal reasoning, we introduce factorised spatiotemporal 3D adapters into the transformer blocks. Existing temporal adapters insert modules \emph{after} self-attention~\cite{liu2024tia} or parallel to the feed-forward network\cite{pan2022stadapter, chen2025msta}, treating temporal modelling as a post-hoc correction to features already computed by the frozen attention layers. Our adapters are placed \emph{before} multi-head self-attention (MHSA) so that the frozen attention operates directly on spatiotemporally enriched tokens, with the temporal signal propagating through the quadratic interaction of the attention softmax. To ensure the adapted tokens are not only temporally coherent but also geometrically structured, we propose a \emph{Token Alignment Loss} that supervises each token position to encode keypoint geometry. For every visible keypoint, a prototype is aggregated from nearby tokens weighted by a Gaussian centred on the ground-truth location, and the loss minimises the KL divergence between the prototype-conditioned cosine-similarity distribution and the Gaussian geometric prior, with a diversity term preventing keypoint collapse. Because the loss operates on the adapter output, gradients flow back through the adapter weights, so the two components are co-trained into a single token-aligned module. We refer to this combination as \emph{TALON}. Our main contributions are:

\begin{itemize}%[noitemsep,topsep=4pt]
\item \textbf{Token Alignment Loss for geometric grounding of visual features.} A novel loss that teaches adapted patch tokens to spatially encode keypoint locations via a prototype-conditioned KL objective against a Gaussian geometric prior, with a diversity regulariser against keypoint collapse. The loss improves pose accuracy and helps preserve domain-agnostic representations for cross-domain generalisation.
\item \textbf{Spatiotemporal Adapter with pre-attention injection.} Factorised 3D adapters placed before self-attention enable temporal modelling in a frozen DINOv3 backbone at less than 5\% additional parameters. Unlike prior designs that insert adapters after attention or parallel to the FFN, our pre-MHSA placement allows the frozen attention to directly reason over temporally enriched tokens, achieving stronger performance with a single adapter per block. We further analyse adapter depth as a control for the specificity/generalisation trade-off.
\item \textbf{Comprehensive multi-domain evaluation} across three spacecraft benchmarks \\ (SPADES\cite{rathinam2024spades}, SPARK\cite{rathinam2024spark}, and SwissCube\cite{hu2021wide}), covering in-domain evaluation on SPADES and SwissCube, in-domain benchmarking on the SPARK real subset, and zero-shot cross-domain transfer from SPADES to SPARK.
\end{itemize}

\section{Related Work}
\label{sec:related}

\subsection{Spacecraft Pose Estimation}
\label{sec:rw_spe}

Monocular 6-DoF spacecraft pose estimation methods fall into two broad categories: direct regression and hybrid keypoint-based approaches~\cite{pauly2023survey}. Direct methods regress the pose vector from the image in a single forward pass~\cite{sharma2018pose, proenca2020deep, phisannupawong2020vision}, but their end-to-end formulation conflates feature extraction and geometric reasoning, limiting interpretability and generalisation across domains. Hybrid methods instead predict intermediate 2D keypoints from a known 3D model and recover the pose through a PnP solver~\cite{chen2019satellite, hu2021wide, park2022speed+}. This decomposition decouples appearance modelling from geometric inference and has become the dominant paradigm, particularly on the SPEED+ benchmark~\cite{park2022speed+}. Chen~\emph{et al.}~\cite{chen2019satellite} advocate high-resolution feature extraction via HRNet for accurate heatmap regression, while Hu~\emph{et al.}~\cite{hu2021wide} address the wide depth-range challenge through multi-scale training. More recently, vision transformers have been adopted as backbone encoders. Wang~\emph{et al.}~\cite{wang2022revisiting} first applied transformers to satellite pose estimation, and FastPose-ViT~\cite{ancey2025fastposevit} demonstrated that a ViT-B backbone with direct regression from the CLS token can be deployed in real time on edge hardware.

A shared limitation across these methods is their reliance on single-frame processing. The few exceptions incorporate temporal information through full fine-tuning of recurrent architectures~\cite{rondao2022chinet, musallam2021leveraging} or auxiliary optical flow networks~\cite{sosa2025motionaware, zuo2024crospace6d}. ChiNet~\cite{rondao2022chinet} appends LSTM units to a CNN backbone, while Sosa~\emph{et al.}~\cite{sosa2025motionaware} combine a ViT encoder with a pre-trained optical flow model \cite{teed2020raft} to inject motion cues. CroSpace6D~\cite{zuo2024crospace6d} similarly relies on explicit flow estimation for cross-domain temporal reasoning. Zhang~\emph{et al.}~\cite{zhang2024monocular} enforce temporal consistency through post-hoc pose smoothing and self-training rather than through the feature representation itself. These approaches either require training all backbone parameters or introduce substantial auxiliary modules, making them poorly suited to the data-scarce spacecraft domain where preserving pre-trained representations is important. Our work differs in that we keep the entire vision backbone frozen and inject temporal reasoning through lightweight adapters that add less than 5\% parameters, while simultaneously grounding the adapted features to keypoint geometry through an explicit alignment objective.

% this section could be shrinked to make space for swisscube
\subsection{Parameter-Efficient Temporal Adaptation}
\label{sec:rw_adapters}

Parameter-efficient transfer learning inserts small trainable modules into a frozen pre-trained model. In the language domain, adapter layers~\cite{houlsby2019parameter} and low-rank updates~\cite{hu2022lora} have become standard practice. For vision transformers, AdaptFormer~\cite{chen2022adaptformer} introduces parallel bottleneck branches alongside the FFN, while VPT~\cite{jia2022visual} prepends learnable tokens to the input sequence.

Extending these ideas to video, ST-Adapter~\cite{pan2022stadapter} inserts depth-wise 3D convolution adapters parallel to the spatial FFN of a frozen CLIP-ViT, factorising spatiotemporal modelling through separate spatial and temporal kernels. AIM~\cite{yang2023aim} further shows that a frozen image encoder can match dedicated video models by inserting lightweight temporal attention and spatial adaptation modules. TIA~\cite{liu2024tia} proposes temporal interaction adapters placed \emph{after} the self-attention layer, modelling inter-frame dependencies as a post-hoc correction to features already processed by the frozen attention. MSTA~\cite{chen2025msta} adopts a similar parallel-to-FFN placement with multi-scale temporal aggregation. A common thread across these methods is their insertion point: adapters operate either after self-attention or parallel to the FFN, meaning the frozen attention mechanism receives unmodified spatial tokens and never directly reasons over temporally enriched representations.

Our adapter design departs from this convention by placing factorised 3D convolution modules \emph{before} multi-head self-attention, allowing the frozen attention to compute queries, keys, and values directly from spatiotemporally enriched tokens rather than applying the temporal signal post hoc. While existing video adapters have been developed exclusively for classification tasks (action recognition), we adapt them to a dense geometric prediction setting (keypoint heatmap regression and pose recovery), which imposes fundamentally different requirements on the spatial structure of the adapted features.

% could be shrinked if needed for Swisscube results 
\subsection{Token-Level Supervision in Vision Transformers}
\label{sec:rw_token}

Self-supervised pre-training objectives such as DINO~\cite{caron2021emerging}, DINOv2~\cite{oquab2024dinov2}, and DINOv3~\cite{Simeoni2025DINOv3} produce patch tokens with strong spatial and semantic properties. Several works have exploited these properties for downstream tasks without modifying them. Amir~\emph{et al.}~\cite{amir2022deep} show that DINO features naturally cluster into semantically meaningful regions, and subsequent work has used frozen DINO tokens for dense correspondence~\cite{caraffa2024freeze}, template matching~\cite{ornek2023foundpose}, and part segmentation. In 6-DoF object pose estimation, FoundPose~\cite{ornek2023foundpose} and Freeze-It~\cite{caraffa2024freeze} leverage frozen DINOv2 features for zero-shot template-based pose retrieval, demonstrating that the representations are geometrically informative even without task-specific training.

However, these approaches treat foundation features as fixed descriptors and do not adapt them to encode a specific geometric structure such as keypoint locations. In the human pose estimation literature, heatmap-based methods~\cite{sun2019deep, nibali2018numerical} supervise the output of a decoder head but leave the backbone features themselves unconstrained with respect to keypoint geometry. Token-level losses have appeared in self-supervised settings: masked image modelling~\cite{he2022masked} reconstructs pixel values from masked tokens, and register tokens~\cite{darcet2024vision} absorb global information to prevent attention artefacts, but neither objective imposes explicit spatial correspondence with downstream geometric targets.

Our patch-token alignment loss addresses this gap. Rather than treating backbone tokens as opaque inputs to a decoder, the loss explicitly supervises the adapted patch tokens to spatially encode keypoint locations through a prototype-conditioned KL-divergence objective. Each keypoint induces a Gaussian geometric prior over the token grid, a local prototype is constructed from nearby tokens, and the loss minimises the divergence between the prototype-conditioned similarity map and the prior. A diversity regulariser prevents multiple keypoints from collapsing onto the same spatial region. Because this loss operates on the adapter output, its gradients co-train the adapter weights, coupling temporal adaptation and geometric grounding into a single module. To our knowledge, no prior work has introduced such a loss for aligning intermediate vision transformer features with geometric keypoint structure in either the general or spacecraft pose estimation literature.

% Spacecraft pose estimation datasets and methods 

% Discuss the robustness of foundation models, and their usage for applied work like pose estimatoin -- modeling geometric vision from visual data

% discuss PEFT methods like adapters and their importance linking to our work 

% discuss token alignment mechanisms linking to our proposed token alignment loss + trajectory smooethness loss 

\section{Methodology}
\label{sec:method}

\begin{figure}[t]
    \begin{center}
    \includegraphics[width=\linewidth]{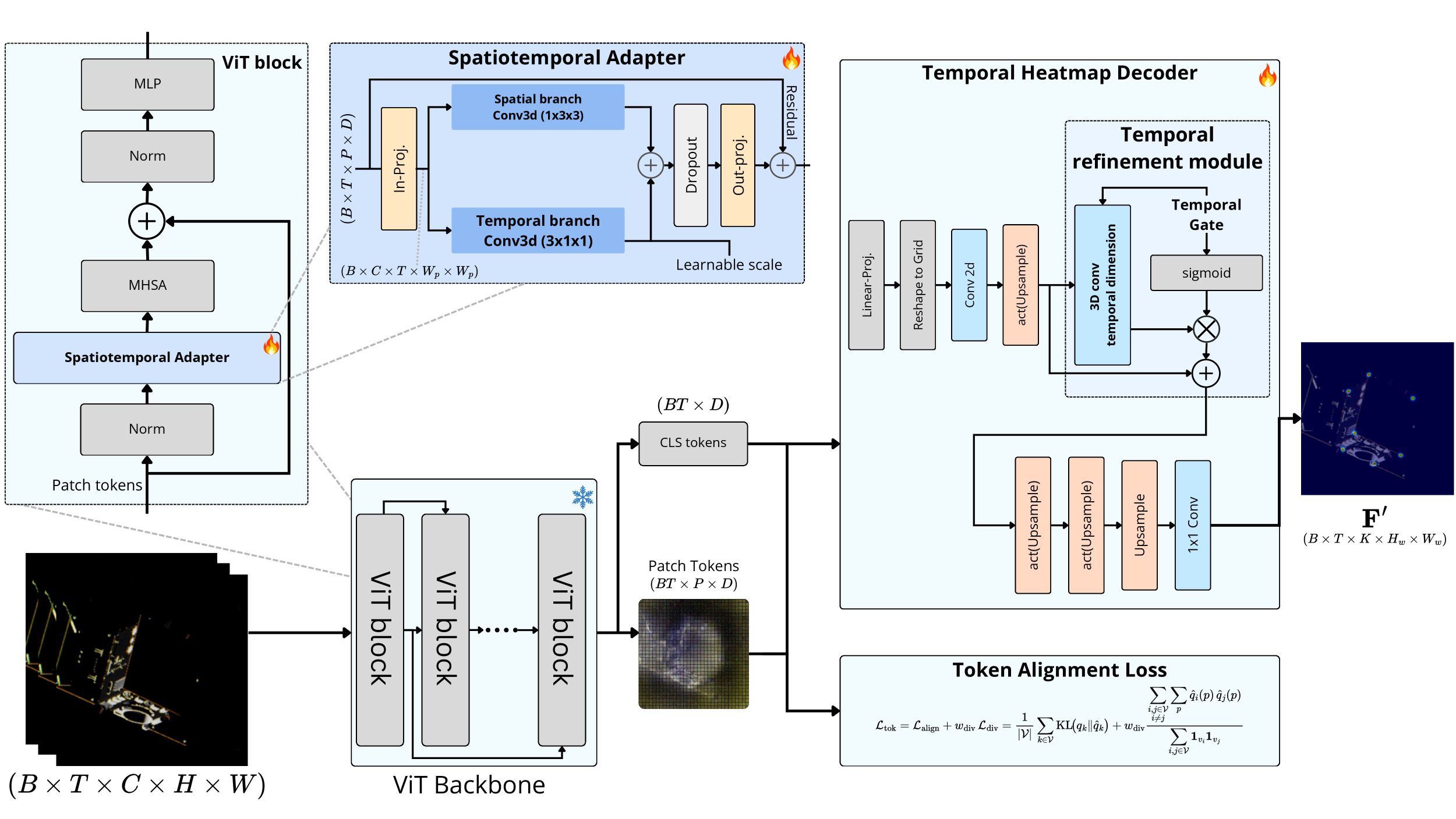}
    \end{center}
    \caption{Overview of TALON. A clip of $T$ frames is processed by a frozen DINOv3 vision transformer; factorised spatiotemporal 3D adapters are injected before multi-head self-attention in the last $L$ transformer blocks (shown in blue). The token alignment loss supervises the adapted patch tokens against Gaussian priors over ground-truth keypoint locations during training. The decoder upsamples the tokens into per-frame keypoint heatmaps, from which DSNT reads out 2D coordinates that are passed to a PnP solver for 6-DoF pose recovery.}
    \label{fig:framework}
\end{figure}
We adapt a frozen vision foundation model to sequential spacecraft imagery through factorised spatiotemporal adapters (\ref{sec:adapters}), a patch-token alignment loss that geometrically grounds the adapted features to keypoint structure (\ref{sec:token_align}), and a temporal heatmap decoder. We refer to the combination of the adapter and the alignment objective as \emph{TALON}. An overview is shown in Fig.~\ref{fig:framework}.

\subsection{Problem Formulation}
\label{sec:formulation}

% Given a temporal clip of $T$ monocular images $\{I_t\}_{t=1}^{T}$ depicting a target spacecraft, the objective is to estimate the 6-DoF pose $(\mathbf{R}_t, \mathbf{t}_t) \in \mathrm{SE}(3)$ for each frame. We follow a keypoint-based pipeline: the model predicts $K$ two-dimensional keypoint coordinates $\hat{\mathbf{k}}_t \in \mathbb{R}^{K \times 2}$ per frame via differentiable spatial-to-numerical transform (DSNT)~\cite{nibali2018numerical}, and the pose is recovered by solving the Perspective-n-Point (PnP) problem using the known 3D keypoint model of the spacecraft. The central challenge is to produce keypoint predictions that are both spatially precise within each frame and temporally coherent across frames. The following subsections describe how each component of our framework contributes to this goal.

Given a temporal clip of $T$ monocular images $\{I_t\}_{t=1}^{T}$ depicting a target spacecraft, the objective is to estimate the 6-DoF pose $(\mathbf{R}_t, \mathbf{t}_t) \in \mathrm{SE}(3)$ for each frame. The model predicts $K$ keypoint coordinates $\hat{\mathbf{k}}_t \in \mathbb{R}^{K \times 2}$ per frame via DSNT~\cite{nibali2018numerical}, and the 6-DoF pose is recovered through PnP solver using the known 3D keypoint model.

\subsection{Spatiotemporal Adapters}
\label{sec:adapters}

The backbone is a frozen DINOv3~\cite{Simeoni2025DINOv3}. To introduce temporal reasoning without altering pre-trained weights, we inject lightweight adapters into the last $L$ blocks, operating exclusively on patch tokens, as illustrated in Fig.~\ref{fig:framework}.

\paragraph{Factorised 3D convolutions.}
Each adapter projects patch tokens through a linear bottleneck $\mathbf{W}_{\mathrm{in}} \in \mathbb{R}^{D \times C}$ and reshapes the result into $(B, C, T, H_p, W_p)$. Two parallel 3D convolution branches process this volume: a spatial branch and a temporal branch, fused by summation:
\begin{equation}
  \mathbf{Z} = Activation\!\bigl(\mathrm{Spatial}_{1 \times 3 \times 3}(\mathbf{Y}) + \mathrm{Temporal}_{3 \times 1 \times 1}(\mathbf{Y})\bigr),
  \label{eq:factorised}
\end{equation}
This factorised design avoids the cubic parameter growth of full $3{\times}3{\times}3$ kernels. The output is projected back via $\mathbf{W}_{\mathrm{out}} \in \mathbb{R}^{C \times D}$ and added through a learnable scale $\alpha$ (initialised to $10^{-3}$):
\begin{equation}
  \hat{\mathbf{x}} = \mathbf{x} + \alpha \cdot \mathbf{W}_{\mathrm{out}} \, \mathrm{Dropout}\!\bigl(\mathbf{Z}\bigr).
  \label{eq:adapter_residual}
\end{equation}

\paragraph{Pre-attention injection.}
Existing temporal adapters operate \emph{after} self-attention~\cite{liu2024tia} or \emph{parallel} to the FFN~\cite{pan2022stadapter, chen2025msta}, adding temporal context as a post-hoc correction to features already processed by frozen attention. We instead place adapters \emph{before} MHSA: the frozen attention receives spatiotemporally enriched tokens as queries, keys, and values simultaneously, and the temporal signal propagates through the quadratic query-key interaction of the softmax. Since the input to MHSA is already layer-normalised, the adapter omits its own LayerNorm. A single adapter per block suffices under this placement; the empirical comparison against post-MHSA injection is reported in Section~\ref{sec:ablation}.

The adapters thus produce patch tokens enriched with temporal context. Temporal enrichment alone, however, does not guarantee that these tokens encode the geometric structure needed for keypoint localisation. A patch token near a spacecraft corner may carry useful appearance and motion features, yet need not activate specifically for the keypoint associated with that corner. The next section introduces the mechanism that provides this geometric grounding.

\subsection{Token Alignment Loss}
\label{sec:token_align}
 
The adapters produce temporally enriched tokens but do not ensure these tokens encode keypoint geometry. Our patch-token alignment loss provides this grounding. For each visible keypoint, it constructs a geometric prior, builds a keypoint prototype from local tokens, and minimises the divergence between the prototype-conditioned similarity map and the prior.
 
\paragraph{Geometric prior.} For keypoint $k$ at ground-truth coordinates $\mathbf{g}_k \in [-1,1]^2$, a target distribution $q_k$ over the $P$ patch positions is defined as a normalised Gaussian:
\begin{equation}
  q_k(p) = \frac{\exp\!\bigl(-\|\mathbf{p}_p - \mathbf{g}_k\|^2 / (2\sigma^2)\bigr)}
               {\sum_{p'}\exp\!\bigl(-\|\mathbf{p}_{p'} - \mathbf{g}_k\|^2 / (2\sigma^2)\bigr)},
  \label{eq:target_dist}
\end{equation}
where $\mathbf{p}_p$ is the spatial coordinate of patch $p$ and $\sigma{=}0.35$.
 
\paragraph{Prototype and predicted distribution.} A keypoint prototype $\boldsymbol{\mu}_k$ is constructed as the $\ell_2$-normalised, $q_k$-weighted sum of $\ell_2$-normalised patch tokens:
\begin{equation}
  \boldsymbol{\mu}_k = \ell_2\!\left(\sum_{p} q_k(p)\,\bar{\mathbf{f}}_p\right),
  \label{eq:prototype}
\end{equation}
where $\bar{\mathbf{f}}_p = \mathbf{f}_p / \|\mathbf{f}_p\|$ and the outer $\ell_2$ normalises the aggregated vector to unit length. This captures the local appearance around $\mathbf{g}_k$ in a geometrically localised subspace. The predicted distribution is the temperature-scaled softmax ($\tau$) of cosine similarities between each token and the prototype:
\begin{equation}
  \hat{q}_k(p) = \frac{\exp\!\bigl(\langle \bar{\mathbf{f}}_p, \boldsymbol{\mu}_k \rangle / \tau\bigr)}
                     {\sum_{p'}\exp\!\bigl(\langle \bar{\mathbf{f}}_{p'}, \boldsymbol{\mu}_k \rangle / \tau\bigr)}.
  \label{eq:pred_dist}
\end{equation}
 
\paragraph{Alignment and diversity.} The loss minimises $\mathrm{KL}(q_k \| \hat{q}_k)$ averaged over visible keypoints $\mathcal{V}$:
\begin{equation}
  \mathcal{L}_{\mathrm{align}} = \frac{1}{|\mathcal{V}|}\sum_{k \in \mathcal{V}} \mathrm{KL}\!\bigl(q_k \| \hat{q}_k\bigr).
  \label{eq:kl_align}
\end{equation}
A diversity regulariser penalises pairwise overlap between the predicted distributions of different visible keypoints:
\begin{equation}
  \mathcal{L}_{\mathrm{div}} = \frac{\displaystyle\sum_{\substack{i,j \in \mathcal{V} \\ i \neq j}} \sum_{p} \hat{q}_i(p)\,\hat{q}_j(p)}{\displaystyle\sum_{i,j \in \mathcal{V}} \mathbf{1}_{v_i}\mathbf{1}_{v_j}},
  \label{eq:div}
\end{equation}
where the denominator counts all visible keypoint pairs (including self-pairs), consistent with the visibility-masked batch normalisation in the implementation. The combined token loss is $\mathcal{L}_{\mathrm{tok}} = \mathcal{L}_{\mathrm{align}} + w_{\mathrm{div}}\,\mathcal{L}_{\mathrm{div}}$. Because $\mathcal{L}_{\mathrm{tok}}$ operates on the adapter output tokens, its gradients flow back through the adapter weights, co-training the two components so that the adapter learns to produce features that are both temporally coherent and geometrically discriminative. The adapter and the alignment loss are therefore co-trained, and the resulting token representations reflect both temporal context and keypoint structure\footnote{Token Alignment Loss is applied only at the training stage}. These token-aligned representations form the input to the decoder described next.

\subsection{Temporal Heatmap Decoder}
\label{sec:decoder}
 
The decoder takes adapted patch tokens $(B, T, P, D)$ and produces per-frame heatmaps at resolution $H_w \times W_w$. It projects features to $C_m$ channels, reshapes to $(H_p, W_p)$, and applies a 2D convolution with GELU followed by upsampling $2{\times}$. At this intermediate resolution, a temporal refinement module re-engages the temporal axis. Just as the backbone adapters model temporal dynamics in the feature space, the decoder's temporal refinement operates in the heatmap space, smoothing and consolidating spatial predictions across frames. It uses a depthwise-separable 3D convolution with kernel size $(k_t, 1, 1)$ along the temporal axis, followed by a pointwise $1{\times}1$ 3D convolution. A sigmoid-gated residual connection controls the contribution of this temporal pathway (as illustrated in Fig.~\ref{fig:framework}):
\begin{equation}
  \mathbf{F}' = \mathbf{F} + \sigma_{\mathrm{gate}}(\gamma) \cdot \mathrm{TemporalRefine}(\mathbf{F}),
  \label{eq:temporal_gate}
\end{equation}
where $\gamma$ is a learnable scalar initialised to zero, mirroring the adapter's $\alpha$ initialisation. Two further upsampling stages and a $1{\times}1$ convolution produce heatmaps $(B, T, K, H_w, W_w)$. Keypoint coordinates are extracted via DSNT~\cite{nibali2018numerical}: spatial softmax followed by expected-value readout yields sub-pixel coordinates.

\section{Experiments}
\label{sec:experiment}

% Dataset description 
\subsection{Datasets.}
% \textbf{Swisscube}, introduced by the EPFL Space Center, is asynthetic dataset for simulating space-borne conditions. It provides 250image sequences of a CubeSat in orbit, captured at varying distances asshown in Fig. 5. Each sequence contains 50 to 100 images, of which100 sequences are allocated to the training set. The images are classified into near, medium, and far categories based on the ratio of thecamera-to-target distance to the spacecraft’s diameter: (\textit{1d, 4d}), (\textit{4d, 7d}), and (\textit{7d, 10d}), where \textit{d}=0.18m denotes the spacecraft diameter.
\textbf{SwissCube}~\cite{hu2021wide} is a synthetic dataset developed by the EPFL Space Center for simulating space-borne imaging conditions. It contains 250 image sequences of a CubeSat in orbit captured at varying distances, with each sequence containing 50 to 100 images. Of these, 100 sequences are used for training and the remainder for testing. Frames are grouped into near, medium, and far categories according to the ratio of camera-to-target distance to the spacecraft diameter $d{=}0.18$\,m, corresponding to the intervals $(1d, 4d)$, $(4d, 7d)$, and $(7d, 10d)$ respectively. We train and evaluate our model on this dataset. 

\textbf{SPADES}~\cite{rathinam2024spades} provides 300 sequences of temporal RGB images depicting a scaled Proba-2 mock-up under diverse lighting and background conditions, captured in Zero-G lab \cite{OLIVARESMENDEZ2023509}. The sequences are divided into 210 training, 45 validation, and 45 test sequences, each containing approximately 600 frames. We use this dataset for training and evaluation.

\textbf{SPARK}~\cite{rathinam2024spark} features the same Proba-2 satellite in two domains: a real subset of 4 sequences (2,048 images) and a synthetic subset of 99 sequences (7,424 images). We use SPARK in two settings: (i) an in-domain setting in which the model is trained and evaluated on the SPARK real subset to compare against published spacecraft pose estimation methods, and (ii) a zero-shot cross-domain setting in which the SPADES-trained checkpoint is evaluated on SPARK without any SPARK supervision.

\subsection{Metrics.}

For SwissCube, we follow the evaluation convention of Hu~\emph{et al.}~\cite{hu2021wide} and report the ADD-0.1d accuracy, which considers a predicted pose correct when the mean 3D distance between model vertices transformed by the predicted and ground-truth poses is below 10\% of the object diameter. We report this accuracy for the near, medium, and far depth regimes as well as the overall score.

For SPADES, and SPARK, we evaluate both 2D keypoint localisation and 6-DoF pose estimation. For keypoint localisation, we report the Percentage of Correct Keypoints (PCK) at thresholds of 1\%, 5\%, and 10\% of the bounding-box diagonal, denoted PCK@1, PCK@5, and PCK@10. For pose estimation, we adopt the standard metrics used in ~\cite{park2022speed+, sosa2025motionaware}: translation error ${E}_{T} = \|\hat{\mathbf{t}} - \mathbf{t}\|_2$, rotation error ${E}_{R} = 2\arccos|\langle\hat{\mathbf{q}},\mathbf{q}\rangle|$, and combined pose score ${E}_{\text{P}} = {E}_{R} + {E}_{T}/\|\mathbf{t}\|$, where $\mathbf{t}$, $\hat{\mathbf{t}}$ are ground-truth and predicted translations and $\mathbf{q}$, $\hat{\mathbf{q}}$ the corresponding quaternions.

\subsection{Results and Discussion}
\paragraph{SwissCube.}
Table~\ref{tab:add_swisscube} reports ADD-0.1d accuracy on the SwissCube test set against SOTA methods. TALON achieves an overall accuracy of 78.8\%, exceeding the previous best method of Chen~\emph{et al.}~\cite{CHEN2026112310} (57.0\%) by 21.8 percentage points, and improving WDR-6D~\cite{hu2021wide} (46.8\%) by 32.0 percentage points. The improvement is consistent across all three distance regimes: 83.1\% at near range, 86.1\% at medium range, and 67.1\% at far range, compared to 65.2\%, 56.9\%, and 47.6\% respectively for Chen~\emph{et al.}

\begin{table}[ht]
\begin{center}
\setlength{\tabcolsep}{8pt}
\begin{tabular}{l|cccc}
\toprule
\textbf{Model} & \textbf{Near [\%]} & \textbf{Medium [\%]} & \textbf{Far [\%]} & \textbf{Overall [\%]} \\
\midrule
SegDriven\cite{hu2019segmentation}      & 41.1 & 22.9 & 7.1  & 21.8 \\
SegDriven-Z\cite{hu2019segmentation}    & 52.6 & 45.4 & 29.4 & 43.2 \\
DLR\cite{chen2019satellite}            & 63.8 & 47.8 & 28.9 & 43.2 \\
WDR-6D\cite{hu2021wide}         & 65.2 & 48.7 & 28.9 & 46.8 \\
Chen\cite{CHEN2026112310}    & 65.2 & 56.9 & 47.6 & 57.0 \\
\rowcolor{gray!10}
\textbf{Ours}  & \textbf{83.1} & \textbf{86.1} & \textbf{67.1} & \textbf{78.8} \\
\bottomrule
\end{tabular}
\end{center}
\caption{Comparison of ADD-0.1d accuracy on SwissCube. Higher is better.}
\label{tab:add_swisscube}
\end{table}

The largest absolute gain of 29.2\% is obtained at medium range, where the spacecraft is sufficiently resolved to expose fine-grained geometric structure but small enough that discriminative keypoint features are difficult to recover from appearance cues alone. The token alignment loss is particularly effective in this regime, since it shapes patch tokens around ground-truth keypoint locations and produces peaked heatmap responses that improve the precision of the PnP fit. The far-range category remains the most challenging at 67.1\%, reflecting the depth ambiguity inherent when the target occupies few pixels. nevertheless, TALON improves on the prior best by 19.5\% in this setting, indicating that the geometric grounding provided by the alignment loss effectively transfers to low-resolution observations. These results support the two central claims of our method: the factorised adapter provides sufficient capacity to specialise a frozen DINOv3 backbone to spacecraft imagery, and the alignment loss produces fine-grained keypoint features that translate into accurate 6-DoF pose recovery across wide depth ranges.

\begin{table}[ht]
\begin{center}
\begin{tabular}{c|ccc|ccc}
\toprule
\multirow{2}{*}{\textbf{Method}} & \multicolumn{3}{c|}{\textbf{2D Keypoint Metrics}} & \multicolumn{3}{c}{\textbf{6-DoF Pose Metrics}} \\
 & \textbf{PCK@10$\uparrow$} & \textbf{PCK@5$\uparrow$} & \textbf{PCK@1$\uparrow$} & $E_T\downarrow$ & $E_R\downarrow$ & $E_P\downarrow$ \\
\midrule
DINOv3~\cite{Simeoni2025DINOv3} & 70.92 & 59.91 & 25.15 & 0.1349 & 0.3527 & 0.4876 \\
Sosa~\cite{sosa2025motionaware}  & 94.05 & 84.05 & 18.17 & 0.0249 & 0.1321 & 0.1570 \\
Ours (ViT-S+)                   & 90.06 & 83.31 & 46.37 & 0.0268 & 0.1472 & 0.1740 \\
\rowcolor{gray!10}
\textbf{Ours (ViT-B)}           & \textbf{94.67} & \textbf{89.79} & \textbf{60.72} & \textbf{0.0177} & \textbf{0.0910} & \textbf{0.1087} \\
\bottomrule
\end{tabular}
\end{center}
\caption{Comparison on SPADES. $E_T$: normalised translation error; $E_R$: rotation error (rad); $E_P = E_R + E_T/\|\mathbf{t}\|$.}
\label{tab:spades}
\end{table}

\paragraph{SPADES.}
Table~\ref{tab:spades} compares our method against two baselines on the SPADES test set: a frozen DINOv3~\cite{Simeoni2025DINOv3} backbone with only the heatmap decoder trained, and the motion-aware ViT of Sosa~\emph{et al.}~\cite{sosa2025motionaware}, which pairs a ViT-B encoder with a pre-trained optical flow model. Our ViT-B variant achieves ${E}_{R}{=}0.0910$\,rad and ${E}_\text{P}{=}0.1087$, reducing the pose score by 31\% over Sosa~\emph{et al.}\ (${E}_\text{P}{=}0.1570$). The improvement is most pronounced at PCK@1, which rises from 18.17\% to 60.72\%. Sub-pixel keypoint accuracy is the precision regime where the token alignment loss has the largest effect, and the $3.3{\times}$ gain at this threshold indicates that the loss shapes the intermediate feature representation to produce sharper, more spatially precise heatmaps. Since the PnP solver is applied deterministically at inference on the predicted 2D keypoints, more accurate keypoint coordinates translate into tighter rotation and translation estimates.
 
The frozen DINOv3 baseline achieves reasonable PCK@10 (70.92\%) but suffers from ${E}_{R}{=}0.3527$\,rad and ${E}_\text{P}{=}0.4876$. Its foundation-model features encode useful spatial structure at a coarse level, yet lack the fine-grained geometric specificity needed for accurate 6-DoF recovery. Despite using a smaller backbone, the ViT-S variant achieves results comparable to those of Sosa~\emph{et al.}\ (${E}_{R}{=}0.1472$\,rad, PCK@10${=}90.06\%$), showing that TALON can exploit even a lower-capacity encoder effectively and remain competitive with methods that employ a ViT-B backbone with full optical flow estimation. Moving to ViT-B yields a higher PCK@1 (60.72\% vs.\ 46.37\%), reflecting the richer feature space available for fine-scale keypoint discrimination.

\begin{table}[ht]
\begin{center}
\setlength{\tabcolsep}{4.3pt}
\begin{tabular}{l| l| ccc| cc}
\toprule
\multirow{2}{*}{\textbf{Method}} & & \multicolumn{3}{c}{\textbf{6-DoF Pose Metrics}} & \multicolumn{2}{c}{\textbf{2D Keypoint Metrics}} \\
 & \textbf{Train} & ${E}_{T}\downarrow$ & ${E}_{R}\downarrow$ & ${E}_\text{P}\downarrow$ & \textbf{PCK@5$\uparrow$} & \textbf{PCK@10$\uparrow$} \\
\midrule
\multicolumn{7}{l}{\textit{Trained on SPARK (in-domain)}} \\[2pt]
CroSpace6D~\cite{zuo2024crospace6d}    & SPARK  & 0.0065 & 0.0187 & 0.0252 & ---   & ---   \\
Liu et al.~\cite{liu2024revisiting}    & SPARK  & 0.0060 & 0.0448 & 0.0508 & ---   & ---   \\
Zhang et al.~\cite{zhang2024monocular} & SPARK  & 0.0092 & 0.0843 & 0.0934 & ---   & ---   \\
\textbf{Ours (Ad2)}                             & SPARK  & \textbf{0.0057} & \textbf{0.0823} & \textbf{0.0924} & \textbf{69.80} & \textbf{79.00} \\
\midrule
\multicolumn{7}{l}{\textit{Zero-shot cross-domain generalisation (trained on SPADES only)}} \\[2pt]
Sosa \cite{sosa2025motionaware} & SPADES & 0.2100 & 0.2670 & 0.3200 & 50.93 & 77.45  \\
\rowcolor{gray!10}
\textbf{Ours (ViT-B, ad2)}             & SPADES & \textbf{0.0332} & \textbf{0.0944} & \textbf{0.1276} & \textbf{65.70} &\textbf{77.33} \\
Ours (ViT-B, ad4)                      & SPADES & 0.0419          & 0.1576          & 0.1995          & 61.10          & 73.10 \\
Ours (ViT-B, ad8)                      & SPADES & 0.0386          & 0.1216          & 0.1602          & 66.40          & 78.80 \\
\bottomrule
\end{tabular}
\end{center}
\caption{Comparison on SPARK real sequences.}
\label{tab:spark_real}
\end{table}

\paragraph{Cross-domain generalisation.}
We evaluate generalisation through zero-shot transfer: models trained on SPADES are applied to SPARK without re-training. On the SPARK real sequences (Table~\ref{tab:spark_real}), the zero-shot Ad2 variant achieves ${E}_P{=}0.1276$, reducing pose error by $2.5{\times}$ relative to Sosa~\emph{et al.}\ (${E}_P{=}0.320$). This is obtained with only 2 adapted blocks, preserving most of the frozen backbone's domain-agnostic features. The Ad8 variant reaches the highest zero-shot PCK@10 (78.8\%) but yields a higher pose error (${E}_P{=}0.1602$) than Ad2. The relationship between adapter depth and cross-domain performance is analysed in Section \ref{sec:ablation}.
 
In the in-domain setting, our Ad2 variant trained on SPARK achieves ${E}_{R}{=}0.0823$\,rad, comparable to Zhang~\emph{et al.}~\cite{zhang2024monocular} (${E}_{R}{=}0.0843$\,rad) despite using a frozen backbone with only two adapter blocks. CroSpace6D~\cite{zuo2024crospace6d} and Liu~\emph{et al.}~\cite{liu2024revisiting} achieve lower translation errors through full model training and auxiliary flow estimation, respectively. These methods do not report keypoint metrics for comparison.
 
On the SPARK synthetic subset (Table~\ref{tab:spark_syn}), the zero-shot Ad2 variant achieves ${E}_P{=}0.0656$, a 34\% reduction over Sosa~\emph{et al.}\ (${E}_P{=}0.0990$), with PCK@1 rising from 19.2\% to 64.1\%. The ViT-S variant maintains competitive results (${E}_P{=}0.1054$, PCK@10${=}94.9\%$) despite its smaller capacity.

\begin{table}[t]
  \begin{center}
  \setlength{\tabcolsep}{3.5pt}
  \begin{tabular}{llccccccc}
    \toprule
    \textbf{Method} & \textbf{Backbone} &
    $E_{T}\downarrow$ &
    $E_{R}\downarrow$ &
    $E_\text{P}\downarrow$ &
    \textbf{PCK@1$\uparrow$} &
    \textbf{PCK@5$\uparrow$} &
    \textbf{PCK@10$\uparrow$} \\
    \midrule
    Sosa et al.~\cite{sosa2025motionaware} & ViT-B & 0.0171 & 0.0819 & 0.0990 & 19.2 & 91.5 & 98.2 \\
    \midrule
    Ours (Ad8)               & ViT-S & 0.0181 & 0.0873 & 0.1054 & 57.1 & 90.6 & 94.9 \\
    Ours (Ad2)               & ViT-B & 0.0114 & 0.0542 & 0.0656 & 64.1 & 93.3 & 97.4 \\
    Ours (Ad4)               & ViT-B & 0.0097 & 0.0433 & 0.0530 & 70.9 & 95.3 & 98.2 \\
    \rowcolor{gray!10}
    \textbf{Ours (Ad8)}      & \textbf{ViT-B} & \textbf{0.0085} & \textbf{0.0397} & \textbf{0.0482} & \textbf{74.9} & \textbf{95.3} & \textbf{98.4} \\
    \bottomrule
  \end{tabular}
  \end{center}
  \caption{SPARK synthetic cross-domain generalisation (trained on SPADES only).}
  \label{tab:spark_syn}
\end{table}

% \paragraph{VBN}
% \paragraph{Cross-domain Evaluation}

\begin{figure}[ht]
    \begin{center}
    \includegraphics[width=\linewidth]{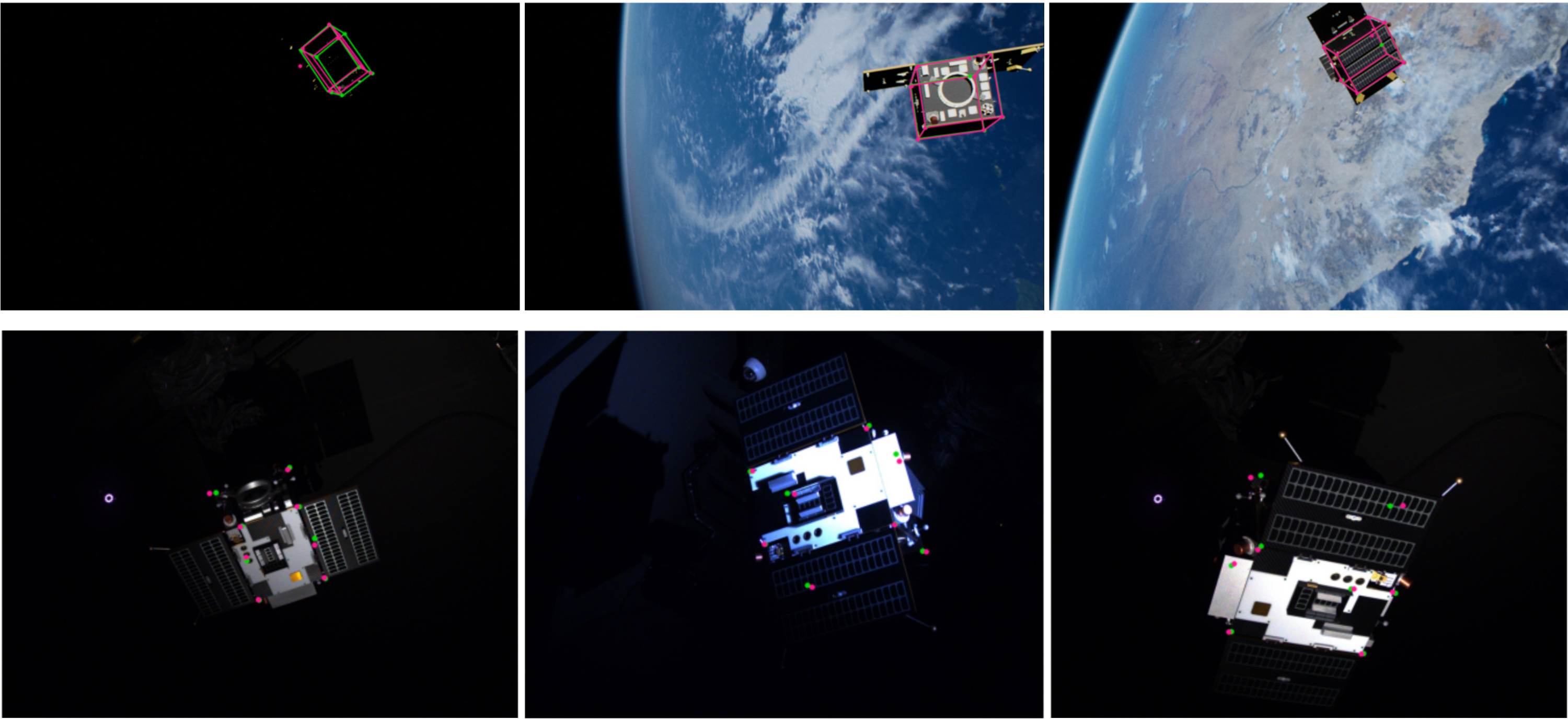}
    \end{center}
    \caption{Qualitative 2D keypoints and projected 3D bounding boxes on SPADES (top) and zero-shot on SPARK (bottom).}
    \label{fig:pose}
\end{figure}

\paragraph{Qualitative analysis.}
Fig.~\ref{fig:pose} shows predicted 2D keypoints and the projected 3D bounding box from the recovered 6-DoF pose on representative frames from SPADES (top row) and SPARK (bottom row). The top row illustrates three test sequences from SPADES spanning far, mid, and close range. Even at far range, where the spacecraft subtends few pixels, the predicted keypoints remain tightly grouped on the object and the projected wireframe aligns closely with the visible geometry. The bottom row shows three zero-shot predictions on SPARK real sequences, where the model has not seen any SPARK data during training. Despite the domain shift (different lighting, Earth-limb background, and rendering pipeline), the keypoints localise accurately on the spacecraft body and the pose wireframe tracks the object geometry consistently.
 
\begin{figure}[ht]
    \begin{center}
    \includegraphics[width=\linewidth]{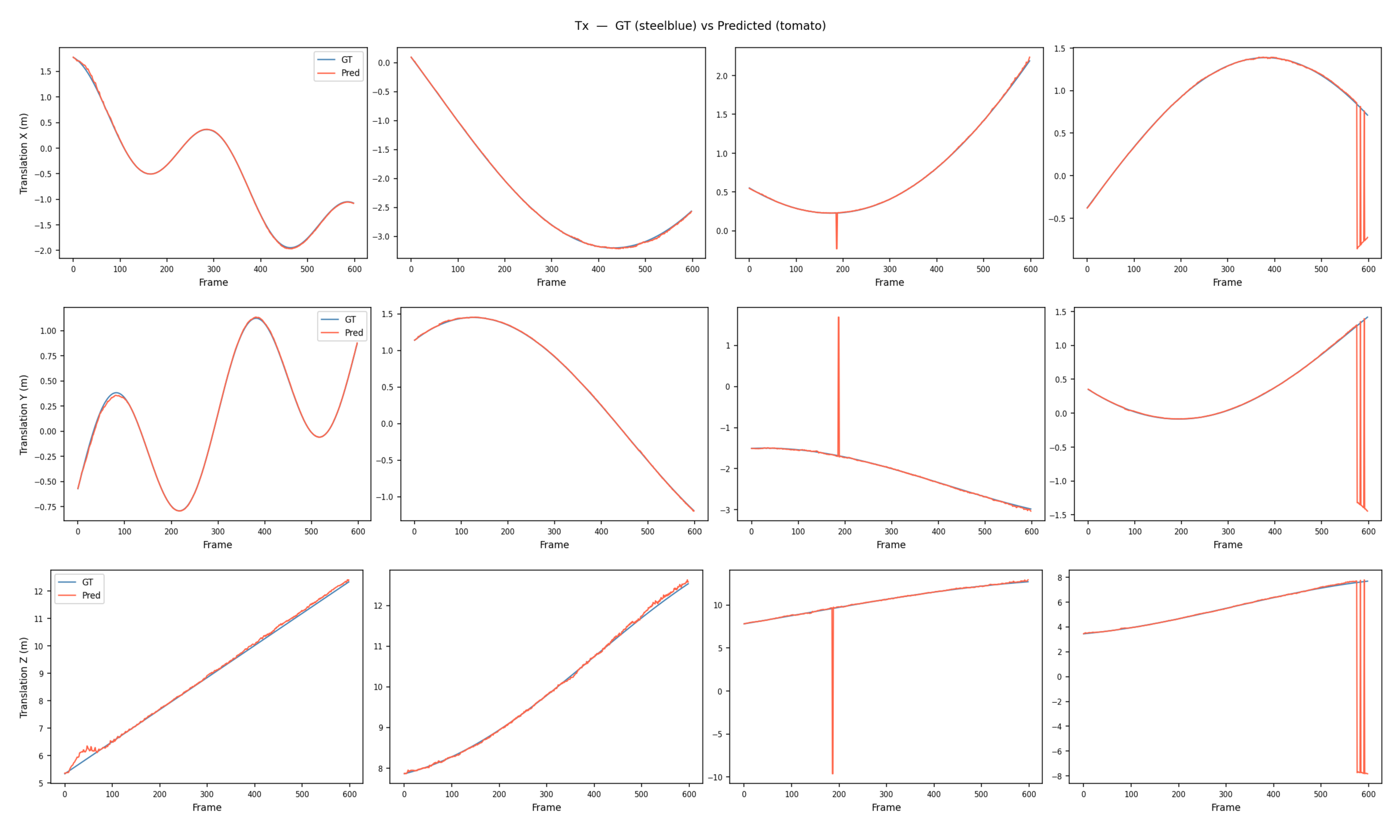}
    \end{center}
    \caption{Per-frame rotation and translation error traces across four SPADES test sequences.}
    \label{fig:traj}
\end{figure}

Fig.~\ref{fig:traj} presents a trajectory smoothness analysis on four randomly selected SPADES test sequences. The rotation and translation error traces over consecutive frames show that the combination of temporal adapters and the decoder's gated temporal refinement produces stable pose trajectories with low frame-to-frame jitter. The error profiles remain smooth across frames, confirming that the temporal modelling components suppress the high-frequency oscillations characteristic of single-frame prediction methods. Occasional error spikes correspond to frames with severe specular reflections or partial occlusions, where even a temporal context cannot fully recover the degraded observation.

\subsection{Ablation Study}
\label{sec:ablation}
 
All ablations use ViT-B/16 on SPADES. Results are in Table~\ref{tab:ablation}.
 
\paragraph{Pre-MHSA vs.\ post-MHSA adapter injection.}
With 8 adapters, token alignment, and cropping fixed, pre-MHSA injection achieves ${E}_P{=}0.1087$ and ${E}_{R}{=}0.0910$\,rad, compared to ${E}_P{=}0.1156$ and ${E}_{R}{=}0.0988$\,rad for post-MHSA. PCK@1 improves from 57.57\% to 60.72\% and PCK@5 from 87.94\% to 89.79\%. The mechanism underlying this difference is described in Section~\ref{sec:adapters}. Empirically, the advantage is preserved while a single adapter per block suffices, keeping the total added parameters below 5\% of the frozen backbone and making the approach suitable for on-board spacecraft deployment, where fine-tuning a full backbone is impractical and model size is limited.

\begin{table}[t]
  \begin{center}
  \setlength{\tabcolsep}{3pt}
  \resizebox{\linewidth}{!}{%
  \begin{tabular}{lcccccccccc}
    \toprule
    \textbf{Injection} & $N_\text{ad}$ & \textbf{Token} & \textbf{Crop} &
    $E_P\!\downarrow$ &
    $E_{T}\!\downarrow$ &
    $E_{R}\!\downarrow$ &
    PCK@1$\uparrow$ &
    PCK@5$\uparrow$ &
    PCK@10$\uparrow$ \\
    \midrule
    Frozen    & 0 & \xmark & \cmark
      & 0.4876 & 0.1349 & 0.3527 & 25.15 & 59.91 & 70.92 \\
    Pre-MHSA  & 8 & \xmark & \xmark
      & 0.2777 & 0.0402 & 0.2375 & 14.48 & 72.39 & 87.39 \\
    Pre-MHSA  & 8 & \cmark & \xmark
      & 0.2580 & 0.0379 & 0.2601 & 26.31 & 73.61 & 88.59 \\
    Pre-MHSA  & 8 & \xmark & \cmark
      & 0.2196 & 0.0203 & 0.1693 & 47.41 & 77.87 & 92.14 \\
    \midrule
    Pre-MHSA  & 2 & \cmark & \cmark
      & 0.1750 & 0.0314 & 0.1436 & 51.71 & 84.40 & 90.50 \\
    Pre-MHSA  & 4 & \cmark & \cmark
      & 0.1190 & 0.0197 & 0.0983 & 56.64 & 87.60 & 93.01 \\
    Post-MHSA & 8 & \cmark & \cmark
      & 0.1156 & 0.0184 & 0.0988 & 57.57 & 87.94 & 93.21 \\
    \rowcolor{gray!10}
    Pre-MHSA  & 8 & \cmark & \cmark
      & \textbf{0.1087} & \textbf{0.0177} & \textbf{0.0910} & \textbf{60.72} & \textbf{89.79} & \textbf{94.67} \\
    \bottomrule
  \end{tabular}%
  }
  \end{center}
    \caption{%
    Ablation on \textbf{SPADES} (ViT-B/16).
    $E_{T}$ (norm) $E_{R}$ (rad) $E_\text{P}$.
  }
  \label{tab:ablation}
\end{table}

\paragraph{Token Alignment and fine-grained geometric precision.}
The ablation isolates the effect of the Token Alignment Loss by comparing two configurations that differ only in the presence of ${L}_{\mathrm{tok}}$: adapters with cropping but no token alignment (${E}_P{=}0.2196$, ${E}_{R}{=}0.1693$\,rad, PCK@1${=}47.41\%$) versus the full model (${E}_P{=}0.1087$, ${E}_{R}{=}0.0910$\,rad, PCK@1${=}60.72\%$). At coarse localisation (PCK@10), the gap is moderate (92.14\% vs.\ 94.67\%). At the strict PCK@1 threshold, the improvement is +13.31 percentage points.
 
PCK@10 measures whether keypoints fall within 10\% of the bounding-box diagonal; at this tolerance, the heatmap and keypoint losses alone suffice for most frames. PCK@1 requires sub-pixel precision, where the alignment loss has the largest effect. By explicitly supervising the adapted patch tokens to spatially encode keypoint locations through the prototype-conditioned KL-divergence objective, the loss shapes the intermediate feature representation, and each keypoint induces a peaked, spatially precise activation pattern in the token field. Sharper heatmaps in turn yield more accurate 2D coordinates under DSNT readout. The reductions in ${E}_{T}$ ($-$13\%) and ${E}_{R}$ ($-$46\%) are a downstream consequence of precise keypoints.
 
The per-sequence analysis corroborates this finding. Token alignment provides the largest improvements on sequences with challenging viewpoints and large range variations (e.g.\ $-$32\% ${E}_P$ on RT232, $-$37\% on RT137, $-$32\% on RT181). On these sequences, the spacecraft appearance changes substantially across frames, and the alignment loss acts as a geometric regulariser that anchors keypoint predictions to the 3D structure of the target. Without it, the adapter features are temporally coherent but not geometrically constrained, and the decoder must infer keypoint geometry from features that carry no explicit spatial correspondence to the 3D model.
 
Cross-domain generalisation also benefits from the alignment loss, which grounds the adapted features to geometric keypoint structure rather than to domain-specific appearance. When the domain shifts (e.g.\ from laboratory to synthetic orbital images), the geometric structure of the keypoints remains constant while textures, lighting, and backgrounds vary. Features aligned with this structure are more transferable than features aligned with appearance alone.

% \paragraph{Adapter depth.}
% In-domain performance scales monotonically with adapter depth. ${E}_{R}$ decreases from $8.23\degree$ ($L{=}2$) to $5.69\degree$ ($L{=}4$) to $5.21\degree$ ($L{=}8$), and PCK@10 increases from 90.5\% to 94.2\% to 94.67\%.
 
% Cross-domain results on SPARK real reveal a different pattern. The shallowest variant ($L{=}2$) achieves the lowest zero-shot pose error (${E}_P{=}0.1276$, ${E}_{R}{=}0.0944$\,rad), outperforming both $L{=}4$ (${E}_P{=}0.1995$) and $L{=}8$ (${E}_P{=}0.1602$). The variant $L{=}8$ recovers in PCK@10 (78.8\% vs.\ 73.1\% for $L{=}4$), suggesting that deeper adapters learn localisation patterns that transfer even when pose-level biases do not. Adapter depth thus acts as a design knob: shallow adaptation preserves domain-agnostic representations for unknown deployment conditions, while deeper adaptation maximises accuracy when training and test distributions align.

\paragraph{Adapters are essential for temporal feature learning.}
The frozen DINOv3 baseline without adapters (${E}_P{=}0.4876$, ${E}_{R}{=}0.3527$\,rad, PCK@10${=}70.92\%$) versus the full model (${E}_P{=}0.1087$, ${E}_{R}{=}0.0910$\,rad, PCK@10${=}94.67\%$) quantifies the contribution of the adapter modules. The $4.5{\times}$ reduction in ${E}_P$ and $3.9{\times}$ reduction in ${E}_{R}$ confirm that the adapters perform the heavy lifting for temporal feature learning. Although the frozen backbone provides strong spatial features, it lacks the inter-frame context needed to resolve transient ambiguities such as specular reflections and motion blur. The factorised 3D adapters introduce this context while keeping all backbone parameters frozen, avoiding catastrophic forgetting of the pre-trained representations.

% \paragraph{Input framing and token alignment.}
% Without bounding-box cropping, the full-image input produces substantially higher errors (${E}_{R}{=}14.43\degree$ without token alignment). The spacecraft occupies a small fraction of the image and most of the tokens correspond to the background. Adding the token alignment loss in this setting reduces ${E}_P$ from $0.295$ to $0.258$ (12.5\% relative). With cropping enabled but without token alignment, ${E}_P$ is $0.231$; adding alignment yields the full model at $0.108$ (53\% relative improvement). The alignment loss thus provides consistent gains across both input regimes, but the effect is larger when cropping concentrates spatial attention on the spacecraft.
 
% \paragraph{Adapter injection position.}
% The placement of Pre-MHSA achieves ${E}_P{=}0.1087$ and ${E}_{R}{=}5.21\degree$, compared to ${E}_P{=}0.126$ and ${E}_{R}{=}5.68\degree$ post-MHSA (both with 8 adapters, token alignment, and cropping). PCK@5 improves from 86.0\% to 89.8\%. This supports the design rationale in Section \ref{sec:adapters}: placing adapters before self-attention lets the frozen attention compute affinities over spatiotemporally enriched tokens, propagating the temporal signal through the quadratic softmax interaction. Pre-MHSA achieves this with a single adapter per block, avoiding the additional parallel branches typical of post-attention designs.

\paragraph{Adapter depth as a generalisation/specificity trade-off.}
In-domain performance on SPADES scales monotonically with adapter depth. In Table~\ref{tab:ablation}, ${E}_{R}$ decreases from $0.1436$\,rad ($L{=}2$) to $0.0983$\,rad ($L{=}4$) to $0.0910$\,rad ($L{=}8$), and PCK@10 rises from 90.50\% to 93.01\% to 94.67\%. Each additional adapted block specialises deeper layers to the training distribution, improving the model's ability to exploit domain-specific visual patterns.
 
However, the cross-domain results on SPARK real in Table~\ref{tab:spark_real} reveal a different pattern. The shallowest variant ($L{=}2$) achieves the lowest zero-shot pose error (${E}_P{=}0.1276$, ${E}_{R}{=}0.0944$\,rad), outperforming both $L{=}4$ (${E}_P{=}0.1995$) and $L{=}8$ (${E}_P{=}0.1602$). Deeper blocks in a vision transformer encode increasingly abstract and domain-specific features. When only the last 2 blocks are adapted, the remaining 10 frozen blocks produce representations that retain the domain-agnostic spatial priors learnt during pre-training, and these priors transfer across domains because they capture general visual structure (edges, textures, spatial relationships) rather than distribution-specific patterns. With 8 adapted blocks, the intermediate representations become tuned to SPADES-specific characteristics (laboratory lighting, background textures, camera noise profile), which do not hold on SPARK.
 
The variant $L{=}8$ recovers partially in PCK@10 (78.80\% vs.\ 73.10\% for $L{=}4$), indicating that deeper adaptation learns keypoint localisation patterns that transfer even when the domain-specific biases in the adapted features degrade pose accuracy. Adapter depth thus functions as a practical design knob: shallow adaptation preserves the frozen backbone's generality for deployment to unknown conditions, while deeper adaptation maximises in-domain accuracy when training and test distributions align.

\paragraph{Input framing.}
Without bounding-box cropping, the full-image input yields substantially higher errors (${E}_{R}{=}0.2375$\,rad, PCK@1${=}14.48\%$; Table~\ref{tab:ablation}). The spacecraft occupies a small fraction of the image, and the majority of patch tokens correspond to background. Adding the token alignment loss in this setting reduces ${E}_P$ from $0.2777$ to $0.2580$ (7.1\% relative). With cropping, the full model achieves ${E}_P{=}0.1087$. The alignment loss provides consistent gains in both regimes, but its effect is amplified when cropping concentrates the spatial attention of the token field on the spacecraft.

\vspace{-13pt}
\section{Conclusion}
\label{sec:conclusion}

We presented TALON for the estimation of monocular 6-DoF spacecraft pose from videos. The approach keeps a frozen DINOv3 vision transformer as the backbone and introduces two complementary mechanisms: factorised spatiotemporal 3D adapters injected before self-attention for temporal reasoning, and a patch-token alignment loss based on KL divergence that geometrically grounds the adapted features to keypoint structure. Pre-MHSA placement allows the frozen attention to compute affinities over spatiotemporally enriched tokens, achieving stronger performance with a single adapter per block than post-attention designs that require additional parallel branches. The alignment loss operates on the adapter output tokens during training, shaping the intermediate representations so that each keypoint induces a peaked, spatially precise activation in the token field. Because the two components are co-trained, the adapter learns to produce features that are simultaneously temporally coherent and geometrically discriminative.

Experiments on SPADES, SPARK, and SwissCube confirm the effectiveness of this design. On SPADES, TALON achieves ${E}_{R}{=}0.0910$\,rad and a $3.3{\times}$ improvement in sub-pixel keypoint accuracy (PCK@1) over existing temporal methods, while adding less than 5\% parameters to the frozen backbone. On SwissCube, the spatial-only configuration attains 78.8\% ADD-0.1d accuracy, surpassing the prior best by 21.8 percentage points and maintaining substantial gains across near, medium, and far depth regimes. Zero-shot transfer to SPARK reduces pose error by $2.5{\times}$ relative to the prior state of the art, and the ablation reveals adapter depth as a control for the trade-off between specificity and generalisation. The token alignment loss contributes most at fine spatial scales and on challenging sequences, acting as a geometric regulariser that anchors predictions to the 3D structure of the target rather than to domain-specific appearance.

Several directions remain open. The current framework assumes a known 3D keypoint model and a bounding-box crop, both of which may be unavailable in unconstrained operational settings. Extending the approach to jointly estimate detection and pose, or to unknown object geometries through category-level priors, would broaden its applicability. The relationship between adapter depth and cross-domain generalisation also warrants further investigation: an adaptive or learned depth selection mechanism could automate the trade-off between specificity and generalisation rather than requiring manual tuning. Finally, integrating the token alignment objective with other dense prediction tasks, such as segmentation or depth estimation, could provide a more complete geometric understanding of the scene beyond keypoint localisation.

\bibliography{egbib}
\end{document}